%% file: eccv2022submission.tex
\documentclass[runningheads]{llncs}
\usepackage{graphicx}
\usepackage[table,xcdraw]{xcolor}
\usepackage{subcaption,booktabs}

\usepackage{tikz}
\usepackage{comment}
\usepackage{amsmath,amssymb} 
\usepackage{color}
\usepackage{graphicx}
\usepackage{booktabs}
\usepackage{paralist}
\usepackage{multirow}
\usepackage{caption}
\usepackage{booktabs} 
\usepackage{algorithm,algcompatible}
\usepackage{scrextend}
\usepackage{dblfloatfix}

\usepackage[pagebackref,breaklinks,colorlinks]{hyperref}

\begin{document}
\pagestyle{headings}
\mainmatter
\def\ECCVSubNumber{8063}  

\title{PseudoAugment: Learning to Use Unlabeled Data for Data Augmentation in Point Clouds}


\titlerunning{PseudoAugment}
%
\author{Zhaoqi Leng$^1$ \and
Shuyang Cheng$^1$ \and
Benjamin Caine$^2$ \and
Weiyue Wang$^1$  \and
Xiao Zhang$^1$ \and
Jonathon Shlens$^2$ \and
Mingxing Tan$^1$ \and
Dragomir Anguelov$^1$
}
\authorrunning{Z. Leng et al.}
%
\institute{
Waymo$^1$, Google$^2$\\
$^1$\{lengzhaoqi, weiyuewang, tanmingxing, dragomir\}@waymo.com
}

\maketitle

\input{abstract}

\input{intro}

\input{relatedworks}
\input{method}
\input{experiment}
\input{conclusion}
\input{acknowledgments}
\clearpage
%
%
\bibliographystyle{splncs04}
\bibliography{egbib}
\end{document}

%% file: abstract.tex
\begin{abstract}
Data augmentation is an important technique to improve data efficiency and save labeling cost for 3D detection in point clouds. Yet, existing augmentation policies have so far been designed to only utilize labeled data, which limits the data diversity. In this paper, we recognize that pseudo labeling and data augmentation are complementary, thus propose to leverage unlabeled data for data augmentation to enrich the training data. In particular, we design three novel pseudo-label based data augmentation policies (PseudoAugments) to fuse both labeled and pseudo-labeled scenes, including frames (PseudoFrame), objects (PseudoBBox), and background (PseudoBackground). PseudoAugments outperforms pseudo labeling by mitigating pseudo labeling errors and generating diverse fused training scenes. We demonstrate PseudoAugments generalize across point-based and voxel-based architectures, different model capacity and both KITTI and Waymo Open Dataset.
To alleviate the cost of hyperparameter tuning and iterative pseudo labeling, we develop a population-based data augmentation framework for 3D detection, named AutoPseudoAugment. Unlike previous works that perform pseudo-labeling offline, our framework performs PseudoAugments and hyperparameter tuning in one shot to reduce computational cost. Experimental results on the large-scale Waymo Open Dataset show our method outperforms state-of-the-art auto data augmentation method (PPBA) and self-training method (pseudo labeling). 
In particular, AutoPseudoAugment is about 3$\times$ and 2$\times$ data efficient on vehicle and pedestrian tasks compared to prior arts. Notably, AutoPseudoAugment nearly matches the full dataset training results, with just 10\% of the labeled run segments on the vehicle detection task. 
\keywords{data augmentation, semi-supervised learning, 3D detection.}
\end{abstract}

%% file: intro.tex
\section{Introduction}
\begin{figure}[t!]
\centering
\includegraphics[ width=0.5\textwidth]{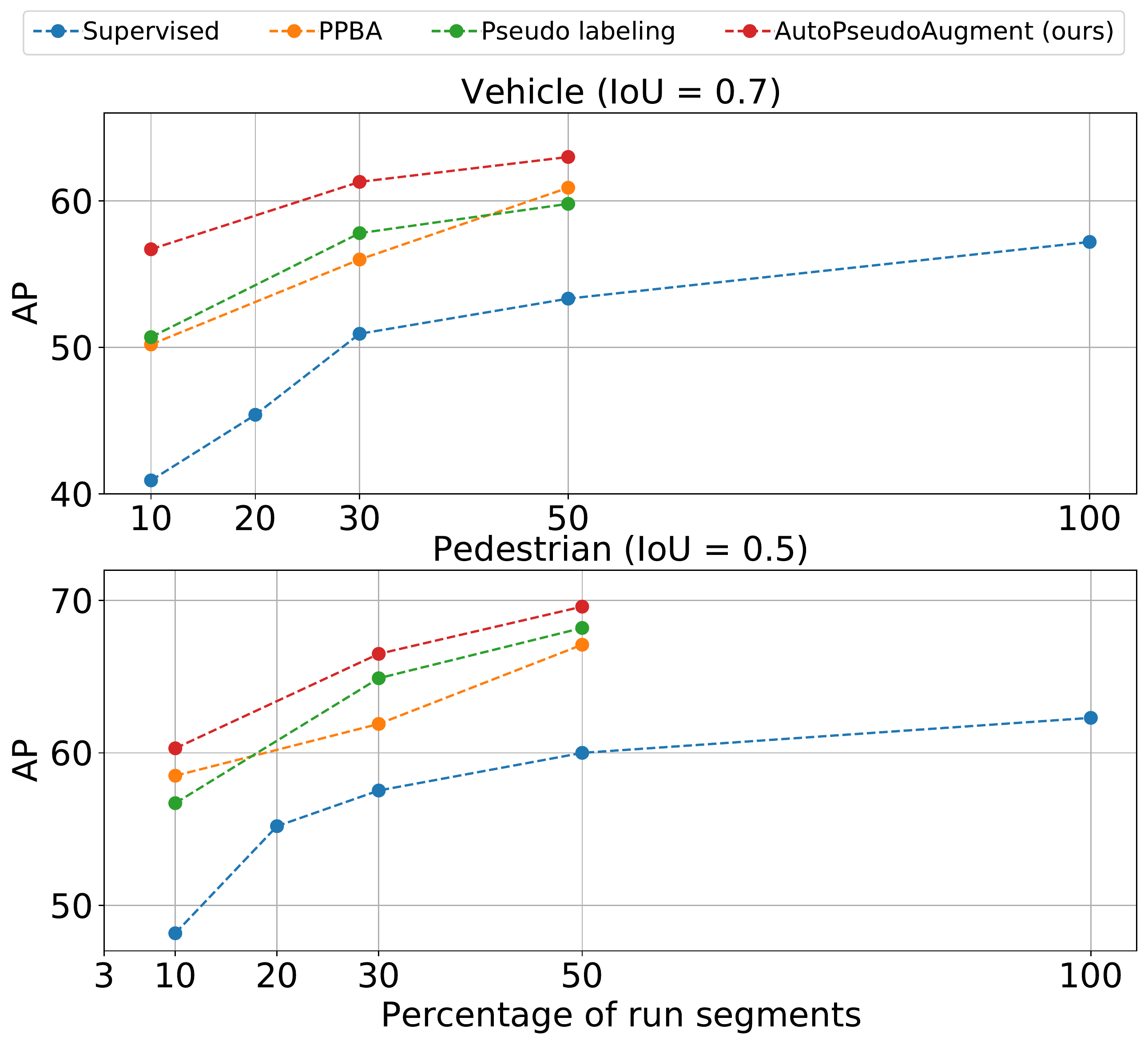}
\caption{\textbf{AutoPseudoAugment is more data efficient than auto data augmentation and self-training methods.}  Data augmentation only (PPBA \cite{cheng2020improving}), self-training (Pseudo labeling \cite{caine2021pseudo}) and our method (AutoPseudoAugment) are evaluated using 3D detection AP at Level 1 difficulty on the \textit{validation} split of the Waymo Open Dataset \cite{sun2020scalability}. Using 10\% of labeled run segments, AutoPseudoAugment is about 3$\times$ data efficient as PPBA and Pseudo label method on the vehicle class and 2$\times$ on the pedestrian class. AutoPseudoAugment is nearly 10$\times$ and more than 5$\times$ data efficient compared to the supervised (no augmentation) vehicle and pedestrian baselines.}
 
\label{fig:efficiency}

\end{figure}

\begin{figure*}[t!]
\begin{center}
\includegraphics[ width=0.70\textwidth]{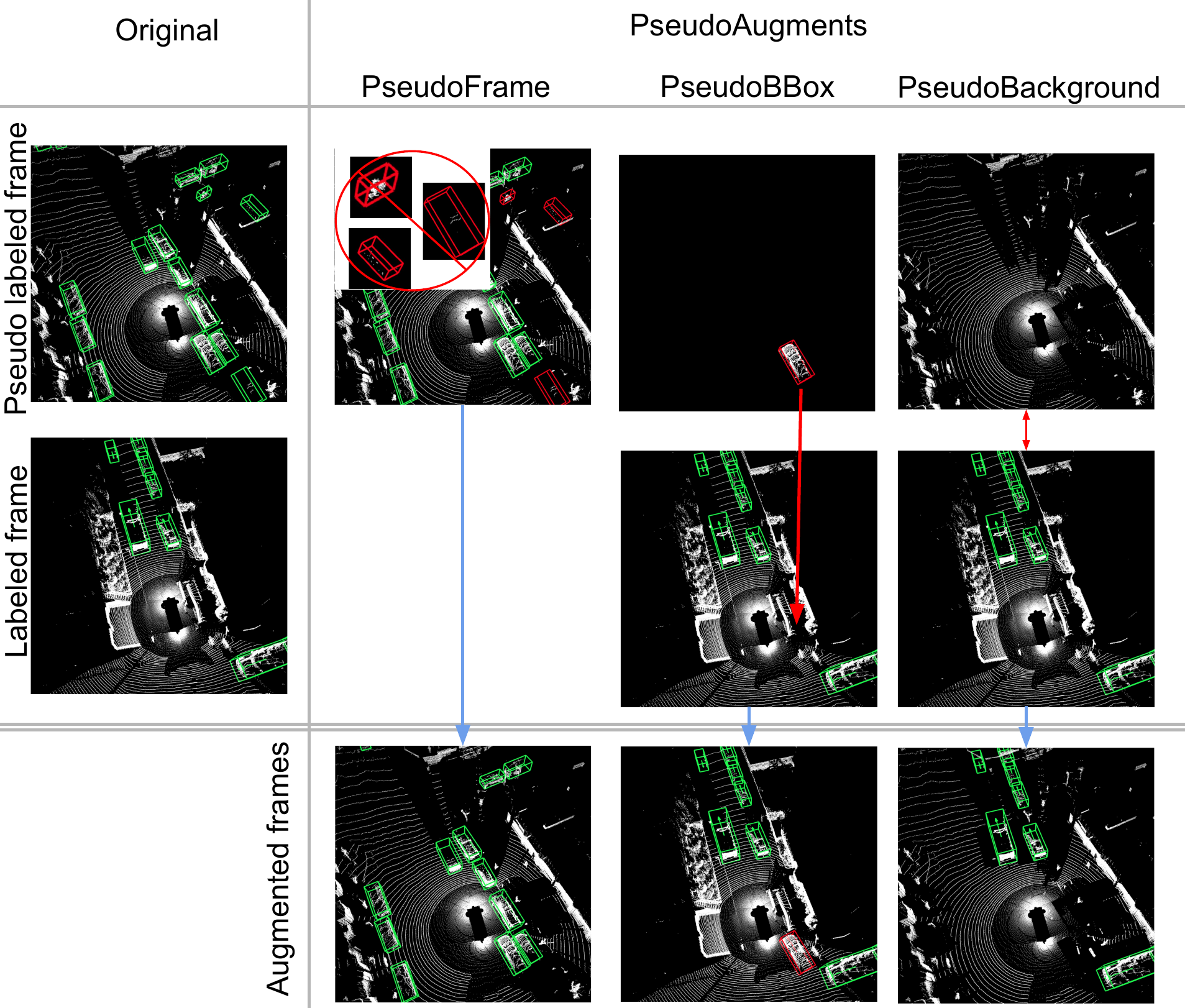}
\end{center}
   \caption{\textbf{Visualization of PseudoAugments}. PseudoAugments contain three new data augmentation policies: PseudoFrame, PseudoBBox, and PseudoBackground. \textbf{PseudoFrame} replaces the labeled frame with a pseudo-labeled frame and drops points of low-confidence bounding boxes in the pseudo-labeled frames. \textbf{PseudoBBox} pastes high-confidence bounding boxes and corresponding point clouds from a pseudo-labeled frame to a labeled frame. \textbf{PseudoBackground} removes all points within bounding boxes in a pseudo-labeled frame, and replaces the background point clouds in the labeled frame with the background point clouds of the pseudo-labeled frame. The augmented frames are used as labeled frames during training. }
\label{fig:pseudoaugment}

\end{figure*}

3D object detection from LiDAR point cloud data is a core component of autonomous driving. Building accurate 3D object detection systems requires vast quantities of labeled scenes with accurate 3D bounding box annotations. While unlabeled LiDAR data is readily available, labeling itself is costly, e.g., 6.4 hours of LiDAR data contains more than 10 million human labeled 3D boxes \cite{sun2020scalability}. Thus, an effective way to increase the data efficiency for model training would be very appealing.

Data augmentation is an effective way to increase data efficiency for labeled data. Data augmentations for 3D detection generally come in two forms: global augmentations like scene rotations, or local augmentations like ground truth augmentation, where crops of ground truth objects from the training set are inserted into the scene. Pasting ground truth objects into the scene has been shown to be extremely effective on various 3D detection datasets \cite{yan2018second,hu2020you,cheng2020improving,yang2018pixor,zhou2018voxelnet,yang2018hdnet,lang2019pointpillars}.

However, these augmentation techniques are typically limited to the labeled training data. A simple way to incorporate unlabeled data into training is pseudo labeling, but naively applying existing 3D data augmentation policies to pseudo labeled frames has an intrinsic limitation, i.e., pseudo labeled frames contain numerous false positive/negative bounding boxes and points. Several recent studies on 3D pseudo labeling \cite{caine2021pseudo,qi2021offboard} have tried to use large-capacity teacher models to mitigate this issue, but the intrinsic pseudo-labeling errors persist. Here, we seek to an alternative approach: \textit{mitigating the pseudo labeled errors by new data augmentation policies}. 

Another challenge is how to effectively combine labeled and unlabeled data via data augmentation. Previous approaches treat pseudo-labeled frames as a whole and do not recognize the compositional nature of 3D point clouds scenes \cite{caine2021pseudo,qi2021offboard}. This limits the diversity of training data. A simple way to fuse labeled and pseudo-labeled frame is to generalize the existing copy-pasting object data augmentation to leverage unlabeled objects. Interestingly, we observe that only pasting objects between labeled and pseudo-labeled frames is not enough \cite{yan2018second,hu2020you,cheng2020improving,yang2018pixor,zhou2018voxelnet,yang2018hdnet,lang2019pointpillars},  because we miss out the diverse background scenes in the pseudo labeled dataset. Especially for 3D point clouds, more than 90\% of the points are backgrounds, which provide critical ingredients for 3D detectors to learn to detect objects in new scenes. Thus, it is necessary to develop a set of data augmentation policies that \textit{take advantage of both foreground objects and background points in the pseudo labeled frames along with labeled frames to generate combinatorial number of point clouds}.

In this work, we propose a set of data augmentation policies tailored for pseudo labeled data, named \textit{PseudoAugments}. As shown in \autoref{fig:pseudoaugment}, our PseudoAugments contain three new data augmentation policies: \emph{PseudoFrame} removes low confidence points, \emph{PseudoBBox} pastes pseudo objects onto labeled scenes, and \emph{PseudoBackground} swaps the background point clouds between labeled and pseudo-labeled scenes. All our augmentations allow pseudo-labeling uncertainty,  and only make use of points of frame, object, and background with high-confidence. PseudoAugments significantly increase the diversity of training data by enabling a combinatorial number of new \textit{fused} training scenes,  including 1) ground truth objects on pseudo labeled background scenes, 2) pseudo labeled objects on ground truth background scenes, and 3) pseudo labeled objects on pseudo labeled background scenes, which greatly enrich the diversity of training data.

Based on PseudoAugments, we develop an auto data augmentation framework named AutoPseudoAugment to learn the best augmentation policies.  Our AutoPseudoAugment is based on population-based training (PBT) and searches for the best augmentation policies online at different training stages. On top of PBT, AutoPseudoAugment uses the top-performing models in previous generations as an ensemble of teachers to pseudo label unlabeled data, which further boosts the quality of pseudo labeled data without the need of training a separated set of high-capacity teacher models \cite{caine2021pseudo,qi2021offboard}. AutoPseudoAugment extends PBT beyond simple hyperparameter tuning by introducing population-based distillation and creates a virtuous cycle between students and teachers, where good teachers in previous generations improves the quality of student models, which become better teachers to pseudo label for future generations.

Our main contributions can be summarized as follows: 
\begin{compactenum}
\item \textbf{PseudoAugments: unifying data augmentation and pseudo labeling.} We identify data augmentation and pseudo labeling are complementary and introduce PseudoFrame, PseudoBBox, PseudoBackground data augmentation policies to take advantage of the composability of unlabeled 3D point clouds while mitigating errors.

\item \textbf{AutoPseudoAugment: efficient one-shot framework for PseudoAugment}. Our framework extends PBT by introducing population-based distillation. AutoPseudoAugment does auto hyperparameters search and self-training in one-shot, which reduces the training cost.

\item \textbf{Extensive experimental evaluations.} We demonstrate PseudoAugments generalize to different network architectures, model sizes, and datasets. In addition, AutoPseudoAugment outperforms both state-of-the-art auto data augmentation method (PPBA \cite{cheng2020improving}) and pseudo labeling \cite{caine2021pseudo}. In particular, leveraging unlabeled data, AutoPseudoAugment requires 10\% of labeled run segments to achieve similar performance as PPBA training on 30\% of run segments and nearly matches the model performance trained on all labeled data without data augmentation, shown in \autoref{fig:efficiency}.

\end{compactenum}

%% file: relatedworks.tex
\section{Related Work}
\subsection{Data augmentation}
Data augmentation has been widely adopted to improve the performance of models trained with supervised learning, such as image classification\cite{simard2003best,ciregan2012multi,wan2013regularization,sato2015apac,krizhevsky2012imagenet,cutout2017,zhang2017mixup}, 2D object detection\cite{girshick2018detectron,dwibedi2017cut}, image segmentation\cite{olaf2015unet,vnet2016,anatomy2015,rosen2018}, point cloud classification and detection\cite{zhou2018voxelnet,yan2018second,chen2017multi,lang2019pointpillars,ngiam2019starnet,shi2019pv,li2020pointaugment,cheng2020improving,fang2021lidar,sheshappanavar2021patchaugment,kim2021point,lee2021regularization,choi2021part,zheng2021se}.

Designing a strong set of augmentation policies for a given task and dataset requires extensive hyperparameter tuning. Automated data augmentation algorithms \cite{lemley2017smart,ratner2017learning,cubuk2018autoaugment,cubuk2020randaugment,lim2019fast,ho2019population} were proposed to search data augmentation policies. Recently, PointAugment \cite{li2020pointaugment} and PPBA \cite{cheng2020improving} introduced automated data augmentation for point clouds, which showed strong empirical results. 

Unlike existing data augmentation methods, which only operates on labeled data, our PseudoAugments are designed to improve the quality of pseudo labeled data and generate combinatorially diverse scenes by fusing labeled and pseudo labeled frames. Different from automated data augmentation frameworks, in particular population-based data augmentation \cite{ho2019population,cheng2020improving}, our AutoPseudoAugment framework enables hyperparameters tuning and self-training in one-shot. It reduces the training cost especially for iterative self-training \cite{xie2020self} and outperforms the state-of-the-art data augmentation framework for 3D point clouds, shown in \autoref{tab:AutoPseudoAugment}.

\subsection{Self-training}
Self-training \cite{mclachlan1975iterative,xie2020self,chen2020semi,caine2021pseudo}, also referred to as pseudo-labeling \cite{lee2013pseudo}, aims to learn from a combination of labeled and unlabeled data. In self-training, a trained teacher network is used to predict labels (pseudo labels) on unlabeled data, and a student model is trained on the combination of the original labeled examples and the new pseudo-labeled examples. Self-training has been applied to a wide variety of tasks, including classification \cite{xie2020self,sohn2020fixmatch,berthelot2019mixmatch}, semantic segmentation \cite{papandreou2015weakly,chen2020semi,wei2016stc,ghiasi2021simple}, object detection \cite{rosenberg2005semi,sohn2020simple,zoph2020rethinking,caine2021pseudo}, speech recognition \cite{park2020improved,kahn2020self}.

Different from prior works on pseudo labeling for 3D point cloud \cite{caine2021pseudo,qi2021offboard,wang20213dioumatch}, where unlabeled frames are used as a whole, our PseudoAugments enables combinatorial new training data by fusing labeled and unlabeled frames. In this work, we aim to demonstrate simple PseudoAugment policies are effective and general methods, while advanced techniques such as IoU-based filtering \cite{wang20213dioumatch}, part\&shape-aware data augmentation \cite{zheng2021se,choi2021part}, and randering-based method \cite{fang2021lidar} could further improve the quality of PseudoAugments.

\subsection{Object Detection for Point Clouds}
There exists a large collection of different architectures for performing 3D object detection. The majority of methods \cite{chen2017multi,yang2018pixor,lang2019pointpillars,zhou2018voxelnet,yan2018second,afdet2020,fan2022embracing} discretize the space into either a 2D (Birds eye view) or 3D grid, and perform either 2D or 3D convolutions on this grid. Some methods alternatively opt to work with the range image view, performing convolutions on the spherical LiDAR image \cite{meyer2019lasernet,bewley2020range,fan2021rangedet}. There also exists a third class of methods, that opt to learn features directly from the raw point cloud \cite{qi2017pointnet,ngiam2019starnet,shi2019pointrcnn,qi2018frustum}, along with a handful of techniques that blend approaches \cite{zhou2020end,shi2019pvrcnn,shi2021pv,sun2021rsn}. Because our method is architecture-agnostic, we view these innovations as complimentary, as our method should benefit current and future architectures.

%% file: method.tex
\section{Methods}
In this section, we first motivate PseudoAugment policies and explain their designs, then we detail the AutoPseudoAugment framework, including our overall data augmentation search process, and how this interacts with PseudoAugment policies. A summary of the algorithm can be found in Algorithm \autoref{algo:autopseudoaugment}.

\begin{algorithm}[h!]
    \begin{algorithmic}
    \scriptsize\STATE {\bf Input}: data and label $(\mathcal{A}, \mathcal{B})$ and unlabeled data $\mathcal{C}$
    \STATE {\bf Init:} set training step $t = 0$, total training steps $\mathcal{N}$, generation step K, randomly initialize M models with random PseudoAugment hyperparameters $\theta$.
    \WHILE{$t \neq \mathcal{N}$}
    \IF{$mod(t, K) == 0$}
        \STATE {\bf \# Population based distillation}
        \STATE {Select the top N models in the previous generation to pseudo label unlabeled data $\mathcal{C}$ and store into a pseudo database which contains unlabeled data and pseudo label $(\mathcal{C}, \mathcal{D})$}
        \STATE {\bf \# Standard progressive PBT exploitation and exploration }
        \STATE {Update hyperparameters $\theta$ and model parameters based on PBT \cite{jaderberg2017population,cheng2020improving} }
    \ELSE
        \STATE{\bf \# Model trained with PseudoAugment policies}
        \STATE {Independently train M models in parallel while using the pseudo database $(\mathcal{C}, \mathcal{D})$ to augment the training data $(\mathcal{A}, \mathcal{B})$ through PseudoAugment policies.}
    \ENDIF
    \ENDWHILE
\end{algorithmic}

\caption{AutoPseudoAugment contains two new elements: population-based distillation and PseudoAugments.}
\label{algo:autopseudoaugment}
\end{algorithm}

\begin{figure*}[!h]
\begin{center}
\includegraphics[ width=0.75\textwidth]{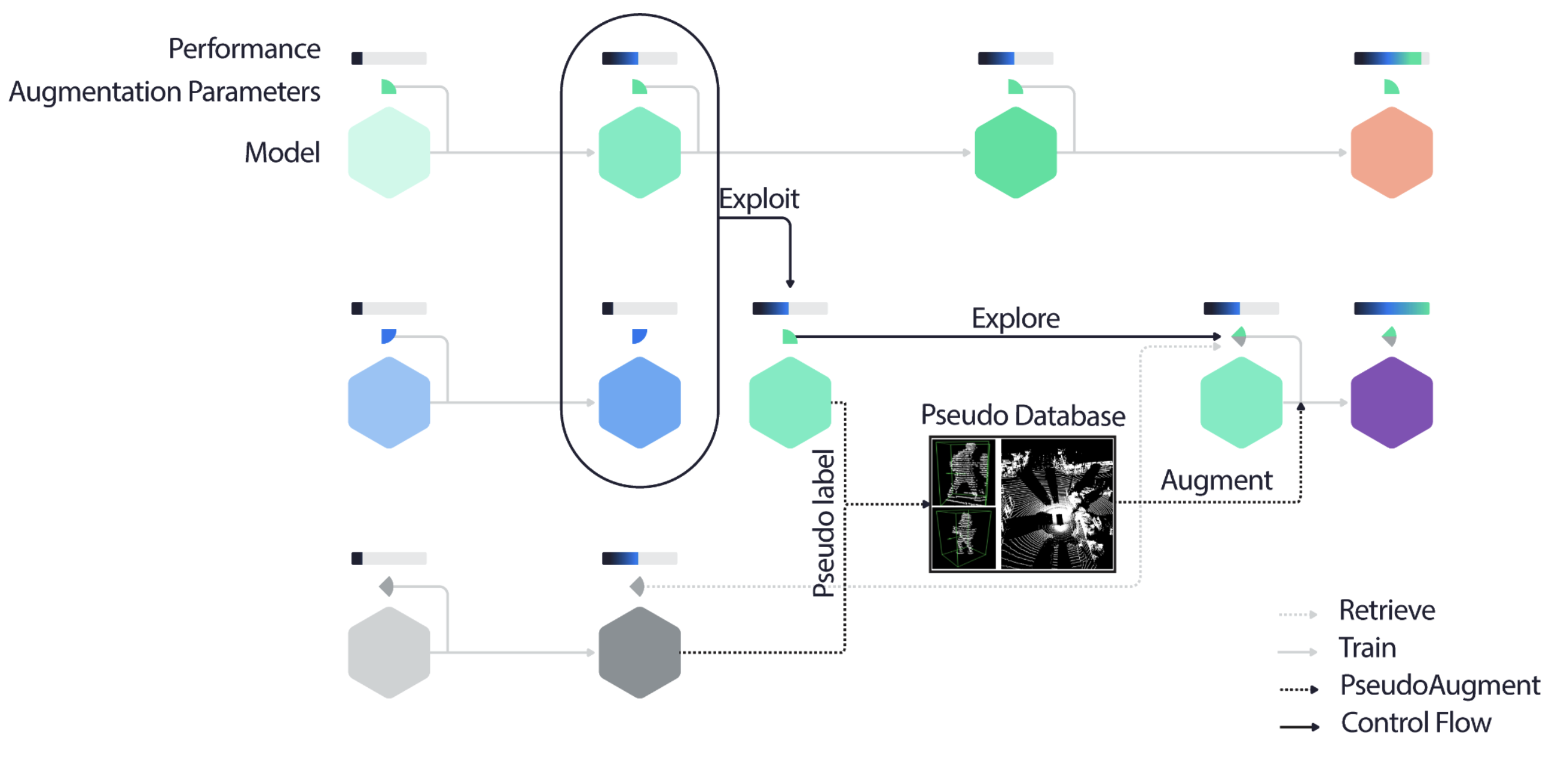}
\end{center}
\caption{\textbf{Schematic diagram of AutoPseudoAugment.} AutoPseudoAugment extends the idea of population-based data augmentation by introducing population-based distillation. Population-based distillation is done at the end of each generation, where we select top-performing models (green and grey hexagons) in previous generations as an ensemble of teachers to pseudo label unlabeled data. The pseudo-labeled frames are stored in a pseudo database, which are used to augment input point clouds in the next generation of training.  In practice, we follow the recommendations of PPBA \cite{cheng2020improving} and only explore up to three augmentation policy choices per generation, with exploration rate 0.8. More details on population-based data augmentation can be found in \cite{ho2019population,cheng2020improving}.}
\label{fig:autopseudoaugment}
\end{figure*}
\subsection{PseudoAugments} \label{sec:pseudoaugments}
The primary goal of PseudoAugments is to generate diverse training data by fusing pseudo-labeled and labeled frames, while reducing mislabeled points and objects in pseudo-labeled frames. We propose three new data augmentation polices which correspond to three different ways of utilizing pseudo-labeled data: PseudoFrame, PseudoBBox, and PseudoBackground.

\textbf{PseudoFrame.} PseudoFrame extends the self-training approach, where a pseudo-labeled frame is used as if a labeled frame during training. Unlike \cite{caine2021pseudo}, where pseudo bounding boxes with low prediction confidence are dropped to reduce false positive bounding boxes, PseudoFrame augments pseudo-labeled frames by removing point clouds within those low confidence pseudo bounding boxes, shown in \autoref{fig:pseudoaugment}. In fact, pseudo labeling is suboptimal compared to PseudoFrame, regardless of what the confidence threshold is, e.g., setting a high threshold will introduce false negative points in the scenes while setting a low threshold will lead to additional false positive pseudo-labeled bounding boxes in the scenes. PseudoFrame mitigates this issue by dropping point clouds in confusing (low confidence) pseudo boxes, which increases the effective quality of pseudo-labeled data, \autoref{fig:precision_recall}, and leads to higher quality student models, \autoref{tab:pseudoaugment}. The PseudoFrame policy contains two hyperparameters: the probability of applying this policy (range [0, 1]), and the threshold of the classification confidence score for both dropping bounding boxes and points (range [0.5, 1]).

Though PseudoFrame leverages unlabeled data and increases the effective quality of pseudo-labeled data, the labeled frames and pseudo-labeled frames are used independently. To further increase the diversity of the training data, we introduce two additional data augmentation policies, i.e., PseudoBBox and PseudoBackground, to fuse labeled frames and pseudo-labeled frames, which introduce a combinatorial number of new training data.

\textbf{PseudoBBox.} Unlike pasting ground truth objects from the labeled frames \cite{yan2018second,cheng2020improving,ngiam2019starnet,lang2019pointpillars}, PseudoBBox is designed to introduce diverse pseudo objects into a training example while reducing the likelihood of pasting false positive points as objects, shown in \autoref{fig:pseudoaugment}. PseudoBBox fuses pseudo-labeled frames and labeled frames by pasting pseudo-labeled foreground objects onto labeled scenes. The PseudoBBox policy contains three parameters: the probability of applying this policy (range [0, 1]), the number of objects that will be added (range [0, 20]), and the threshold of the classification confidence score (range [0.5, 1]) required for an object to be inserted into a scene. 

To align pasted objects to the new background scene, we adjust the $Z$ coordinate of pasted objects based on the estimated $Z$ coordinate of the new ground plane \footnote{We estimate the ground plane using linear regression of the bottom center of all the foreground ground truth or pseudo-labeled objects. If a scene contains less than 3 bounding boxes, we use the histogram of point clouds Z coordinate to estimate the ground plane.}. We oversample 10$\times$ pseudo objects and reject pseudo objects that overlap with any other pseudo objects and existing ground truth objects in the scene, then sample from the reminding pseudo objects and paste the predefined number of pseudo objects into the scene. If the pasted objects overlap with background points, we will remove background points. 

\textbf{PseudoBackground.}  Perhaps surprisingly, the background point clouds in unlabeled data contain important ingredients for generating diverse fused scenes, which were not recognized. Simply swapping the background point clouds in labeled frames and unlabeled frames, we generate diverse fused training scenes with ground truth objects on top of background point clouds from pseudo-labeled scenes. Different from PseudoFrame and PseudoBBox, we aggressively reject both true negative and false negative points in point clouds by removing all points within pseudo bounding boxes with object classification confidence scores above 0.1, and use reminding points as the background point clouds. Thus, the PseudoBackground contains only one hyperparameter, i.e., the probability of applying this operator (range [0, 1]). We align the ground plane of the new pseudo background point cloud with the existing point cloud and reject pseudo background point clouds when overlapping with bounding boxes, following the process described above for PseudoBBox.

\subsection{AutoPseudoAugment}
AutoPseudoAugment is a data augmentation framework designed for efficient hyperparameter tuning while applying PseudoAugments in one shot, shown in Algorithm \autoref{algo:autopseudoaugment}.

\textbf{Population-based distillation} Motivated by the recent success of population-based augmentation \cite{ho2019population,cheng2020improving}, we use PBT to search hyperparameters in PseudoAugments. However, traditionally, hyperparameter tuning and self-training are decoupled. Especially for iterative self-training \cite{xie2020self}, tuning the hyperarameters for the student model in each iteration will incur significant computation cost. To mitigate this challenge, we propose population-based distillation, where we take advantage of the models in previous generations as an ensemble of teachers to pseudo label unlabeled data, shown in \autoref{fig:autopseudoaugment}.

Unlike PBT, where model checkpoints in previous generations are discarded when training models, we recycle and use the top N model checkpoints in the previous generation as teachers. Because the previous generation checkpoints are trained with different schedules of data augmentation policies, they naturally form a diverse set of teachers. Thus, population-based distillation achieves both hyperparameter tuning and ensemble distillation at once.

In addition to our three new PseudoAugment policies PseudoFrame, PseudoBBox, and PseudoBackground, we adopt the full suite of data augmentations used by PPBA \cite{cheng2020improving}. In order to further increase the diversity of our training data, we apply our three PseudoAugment policies \textit{before} we apply other geometric-based data augmentations, allowing pseudo-label augmented scenes to be further augmented by other common data augmentations. Our final order of augmentations are: PseudoFrame, PseudoBoundingBox, PseudoBackground, RandomRotation, WorldScaling, GlobalTranslateNoise, FrustumDropout, FrustumNoise, RandomDropLaserPoints. 

%% file: experiment.tex
\section{Experiments}
We extensively evaluate PseudoAugments policies and AutoPseudoAugment framework using voxel-based PointPillars model \footnote{Code for both models are available at https://github.com/tensorflow/ lingvo/tree/master/lingvo/tasks/car under Apache License 2.0.\label{lingvo}} \cite{lang2019pointpillars} and point-based StarNet model \footref{lingvo} \cite{ngiam2019starnet} on KITTI \cite{kitti2013} and Waymo Open Dataset \cite{sun2020scalability}. 

For the following experiments, we train two separate models to detect vehicles and pedestrians and adopt the same training setting as prior works \cite{ngiam2019starnet,cheng2020improving,caine2021pseudo}. To study the data efficiency, we create a smaller training set consisting of 10\%, 30\% and 50\% of the run segments from the Waymo Open Dataset training set to use as our labeled dataset, while using the remaining run segments as an unlabeled dataset. We want to highlight that 10\% of the Waymo Open Dataset contains a considerable amount of 3D labeled bounding boxes (more than 1 million) which is on par with other full training dataset such as KITTI, NuScenes, and Argoverse dataset \cite{geiger2013vision,caesar1903nuscenes,chang2019argoverse}. For hyperparameter tuning on Waymo Open Dataset, we create a random subsampling of the validation set, using 10\% of examples (4109 samples) as \textit{mini-val} and use Level 1 difficulty average precision (AP) as our objective value. 

\begin{table*}[!t]%
        \centering
        \resizebox{0.85\textwidth}{!}{
            \begin{tabular}{ll|l|l}
            \toprule
            Setup & Effects & Vehicle L1  AP & Pedestrian L1 AP \\ \midrule
            Supervised  (Teacher) &  & 49.6 & 53.9 \\ \midrule
            Pseudo labeling  \cite{caine2021pseudo}&  & 50.7 \color[HTML]{3166FF} (+1.1)& 56.7 \color[HTML]{3166FF} (+2.8)\\ 
            PseudoFrame only (Ours)& Reducing errors & 51.1 \color[HTML]{3166FF} (+1.5) & 57.2 \color[HTML]{3166FF} (+3.3) \\ 
            PseudoBBox only (Ours)& Fusing scenes & 53.4 \color[HTML]{3166FF} (+3.8) & 57.0 \color[HTML]{3166FF} (+3.1) \\ 
            PseudoBackground only (Ours)& Fusing scenes & 51.9 \color[HTML]{3166FF} (+2.3) & 57.7 \color[HTML]{3166FF} (+3.8) \\ 
            All PseudoAugments (Ours)& Reducing errors + fusing scenes & {\textbf{54.3} \color[HTML]{3166FF} \textbf{(+4.7)}} & \textbf{58.4} {\color[HTML]{3166FF} (\textbf{+4.5})} \\ \bottomrule
            \end{tabular}
        }

     \caption{\textbf{PseudoAugments improve upon Pseudo labeling method}. PseudoAugments reduce errors in the pseudo-labeled scenes by dropping low-confidence points, and improves data diversity by introducing fused pseudo-labeled scenes. Supervised PointPillars models are trained on 10\% run segments and used as teachers. Pseudo labeling drops pseudo-labeled bounding boxes below confidence threshold 0.5, while PseudoFrame augments pseudo-labeled scenes by dropping both bouding boxes and point clouds within those bounding boxes below threshold 0.5. The improvements from PseudoAugments are additive. Introducing PseudoBBox and PseudoBackground further enrichs the training data, which leads to better student models. 3D detection Level 1 AP are evaluated on the Waymo Open Dataset \textit{validation set}. }
     \label{tab:pseudoaugment}
\end{table*}

\subsection{PseudoAugments helps quality and diversity} \label{sec:quality_diversity}
In this section, we show PseudoAugments reduce the errors in pseudo labeled scenes via PseudoFrame and can generate diverse fused scenes when applying PseudoBBox and PseudoBackground, which outperform pseudo labeling method for both vehicle and pedestrian detection tasks. We follow the implementation in \cite{caine2021pseudo} and train teacher models on 10\% of the training run segments using random Z rotation and random flip Y data augmentation for 150 epochs. We use the teacher models to pseudo label the reminding 90\% of the training run segments and remove pseudo-labeled bounding boxes with classification score below 0.5. When training the student models, we use 1:1 ratio of labeled and pseudo labeled scene in each mini batch. Since the training data is increased 10$\times$, we train the student model for 10$\times$ steps to take advantage of the additional pseudo labeled data, results shown in \autoref{tab:pseudoaugment}.
\begin{figure}[!h]
     \centering
     \includegraphics[width=0.45\textwidth]{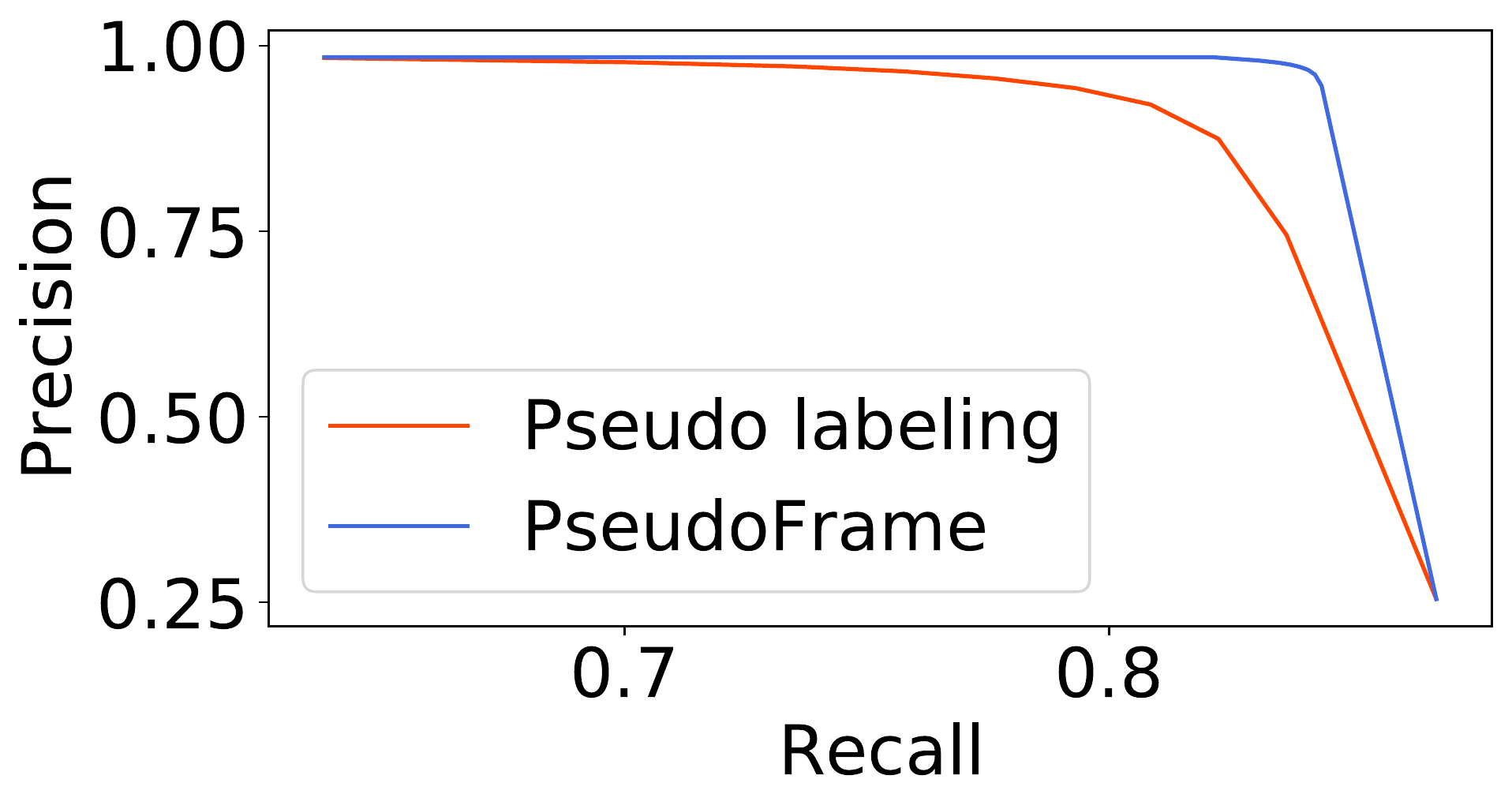}
    \caption{\textbf{PseudoFrame improves quality of pseudo labeled point clouds.} Precision and recall are defined based on whether a point is inside labeled/pseudo labeled vehicle or pedestrian bounding boxes. Vanilla pseudo labeling approach only adds pseudo bounding boxes if the prediction confidence is higher than 0.5, but keeps all the false-negative points; In contrast, our PseudoFrame also drops points in low-confidence bounding boxes, thus reducing false negatives and improving precision-recall of pseudo labeled frames.}
\label{fig:precision_recall}
\end{figure}

\textbf{PseudoFrame improves data quality.} PseudoFrame augments the pseudo labeled scenes by removing point clouds in low-confidence pseudo bounding boxes. Here, we remove bounding boxes and corresponding point clouds with classification confidence score below 0.5. As shown in \autoref{fig:precision_recall}, simply removing those point clouds is an effective data augmentation to increase the prevision-recall of pseudo labeled points. PseudoFrame improves the quality of student models (+0.4 on Vehicle AP and +0.5 on Pedestrian AP) compared to Pseudo labeling, shown in \autoref{tab:pseudoaugment}.

\textbf{PseudoBBox and PseudoBackground increase diversity.}
PseudoBBox and PseudoBackground increase the diversity of the training scenes by fusing pseudo labeled and labeled scenes, as shown in \autoref{fig:pseudoaugment}. To find the optimal hyperparameters, we randomly sample 16 different combinations of hyperparameters from the search space detailed in \autoref{sec:pseudoaugments}. Introducing fused scenes further improves the quality of student models (+3.2 on Vehicle AP and + 1.2 on Pedestrian AP) compared to only applying PseudoFrame data augmentation, \autoref{tab:pseudoaugment}, which shows the benefit of PseudoBBox and PseudoBackground is additive.

\subsection{Generalization of PseudoAugments}\label{sec:generalize}

In the previous section, we demonstrate PseudoAugments improves upon pseudo labeling method on PointPillars models. In this section, we show PsueodAugments generalizes to different model sizes and architectures. In addition to PointPillars model, which is a voxel-based architecture \cite{lang2019pointpillars}, we evaluate PseudoAugment on larger PointPillars models and point-based StarNet \cite{ngiam2019starnet} models. We use the same pseudo labeled data as in \autoref{sec:quality_diversity}, which is labeled by the supervised PointPillars models shown in \autoref{tab:pseudoaugment}. We show besides self-training using the same model size and architectures, PseudoAugments enables self-training from a smaller model to a larger model and across different architectures. Our results show PseudoAugments lead to higher improvements compared to pseudo labeling, \autoref{tab:generalize}. For the following experiments, we adopt the same training settings as in \autoref{sec:pseudoaugments}.

\begin{table*}%
     \centering
    \begin{subtable}[b]{0.45\textwidth}
        \centering
        \resizebox{\textwidth}{!}{
            \begin{tabular}{l|l|l}
            \toprule
            Setup & Vehicle  AP & Pedestrian AP \\ \midrule
            Supervised & 52.1 & 56.9 \\ \midrule
            Pseudo labeling \cite{caine2021pseudo}& 51.6 \color[HTML]{FF0000} (-0.5)& 57.8 \color[HTML]{3166FF} (+0.9)\\ 
            All PseudoAugments (Ours) & \textbf{55.7}{\color[HTML]{3166FF} (\textbf{+5.5})} & \textbf{59.7} {\color[HTML]{3166FF} (\textbf{+2.8})} \\ \bottomrule
            \end{tabular}
        }
    \caption{Pillars2X.}
    \label{tab:pillars2X}
     \end{subtable}
         \quad
    \begin{subtable}[b]{0.45\textwidth}
        \centering
        \resizebox{\textwidth}{!}{
            \begin{tabular}{l|l|l}
            \toprule 
            Setup & Vehicle  AP & Pedestrian AP \\ \midrule
            Supervised & 43.7 & 60.6 \\ \midrule
            Pseudo labeling  \cite{caine2021pseudo}& 48.2 \color[HTML]{3166FF} (+4.5) & 63.5 \color[HTML]{3166FF} (+2.9)\\ 
            All PseudoAugments (Ours) & \textbf{51.2} {\color[HTML]{3166FF} (\textbf{+7.5})} & \textbf{64.7} {\color[HTML]{3166FF} (\textbf{+4.1})} \\ \bottomrule
            \end{tabular}
        }
    \caption{StarNet models.}
    \label{tab:starnet}

     \end{subtable}
     \caption{\textbf{PseudoAugments generalize to larger capacity models and different architectures.} PseudoAugments outperform pseudo labeling on 10\% run segments using PointPillars \cite{lang2019pointpillars}, in \autoref{tab:pseudoaugment}, as teachers. (a) self-training from PointPillars teachers to larger PointPillars models (Pillars2X) and (b) self-training from PointPillars teachers to StarNet models \cite{ngiam2019starnet}. Note that PseudoAugments improve the vehicle detection quality of Pillars2X whereas pseudo labeling is unable to.  3D detection Level 1 AP are evaluated on the Waymo Open Dataset \textit{validation set}. }
   \label{tab:generalize}
\end{table*}

\textbf{Generalize to larger models.}
We double the channel numbers of every convolution layers in the PointPillars model and denote it as Pillars2X. We train Pillars2X on the same supervised 10\% run segments as the supervised training baseline, which has higher quality compared to the standard (1x) PointPillars, shown in \autoref{tab:pillars2X}. Interestingly, the pseudo labeling method failed to improve the vehicle Pillars2X model (52.1 AP) when we use a weaker (1X) model (49.6 AP) as teacher. This indicates errors in pseudo labeled frames diminishe the benefit of introducing unseen scenes to diversify the training data. Whereas, applying PseudoAugments overcomes this limitation and leads to significant improvement (+4.1 on Vehicle AP and + 1.9 on Pedestrian AP) compared to pseudo labeling.   

\textbf{Generalize to different architectures.}
Unlike voxel-based PointPillars, StarNet is a point-based 3D detector and learns feature representations directly from raw point clouds.  Our results show, using PointPillars model as teacher, PseudoAugments significantly improves quality of the StarNet student models (+3.0 on Vehicle AP and + 1.2 on Pedestrian AP) compared to pseudo labeling method \autoref{tab:starnet}. This shows PseudoAugments are model agnostic and outperform pseudo labeling method when we distill two models with very different architectures.

\subsection{Generalize to KITTI dataset}
In this section, we show PseudoAugments is a general method that is effetive on significantly different datasets. Different from Waymo Open Dataset \cite{sun2020scalability}, KITTI \cite{kitti2013} dataset was collected in different cities and has different point and object density per frame. Here, we follow the common practice and split the KITTI dataset in half, i.e., one used for training and the other half used for validation. We randomly select 10\% of the training frames as a mini training split, while removing labels on the rest 90\% of the training frames. We train PointPillars teacher models on the mini training split with random flip and random world scaling data augmentations. Our results, in \autoref{tab:kitti}, show using PseudoAugments consistently outperform pseudo labeling on detecting objects at all difficulties. 
\begin{table*}[!h]%
        \centering
        \resizebox{0.55\textwidth}{!}{
            \begin{tabular}{l|l|l}
            \toprule
            Setup &  Vehicle (E/M/H) & Ped\&Cyc (E/M/H) \\ \midrule
            Supervised  (Teacher) & 55.6/49.2/46.1 & 46.3/33.4/30.2 \\ \midrule
            Pseudo labeling  \cite{caine2021pseudo}&  64.3/51.4/49.0 & 49.3/35.9/32.6\\ 
            All PseudoAugments (Ours)& \textbf{65.5/56.5/53.7} & \textbf{55.2/40.8/37.5} \\ \bottomrule
            \end{tabular}
        }
     \caption{\textbf{PseudoAugments generalize to KITTI dataset}. PseudoAugments outperform pseudo labeling on 10\% KITTI training frames using PointPillars \cite{lang2019pointpillars} as teachers. 3D detection APs for easy, moderate, and hard (E/M/H) objects are evaluated on the KITTI \textit{validation set}. }
     \label{tab:kitti}
\end{table*}

\subsection{AutoPseudoAugment improves data efficiency} 
In previous sections, we demonstrate PseudoAugments are strong data augmentation methods that improves upon pseudo labeling. In this section, we show AutoPseudoAugment leverages PseudoAugments and further advances state-of-the-art auto data augmentation methods for 3D point clouds (PPBA) \cite{cheng2020improving}.

When the models are trained on 10\% labeled run segments, we use generation step 1000 for both PPBA and AutoPseudoAugment. On 30\% and 50\% run segments, we increase the generation step to 2000. Even though AutoPseudoAugment introduces additional PseudoAugment policies with more hyperparameters compared to PPBA, we use the same number of tuners (population size 16) for AutoPseudoAugment and PPBA. We follow the other training settings in \cite{cheng2020improving}. At the end of each generation, we select the top 10 models in previous generations with L1 AP above 0.35 as ensemble of teachers to pseudo label unlabeled data. 

\textbf{AutoPseudoAugment outperforms both Pseudo labeling and PPBA methods} Our AutoPseudoAugment framework subsumes both the auto data augmentation and pseudo labeling, which takes advantage of additional unlabeled data while tuning hyperparameters online.  More importantly, PseudoAugments generate high-quality fused scenes, which greatly increases the diversity of the training data. As shown in \autoref{tab:AutoPseudoAugment}, AutoPseudoAugment outperforms both PPBA and Pseudo labeling on 10\%, 30\%, and 50\% labeled run segments. 

To estimate the data efficiency, we train PointPillars models without data augmentation on 10\%, 20\%, 30\%, 50\% and 100\% of training run segments, shown in \autoref{fig:efficiency}. According to this metric, our AutoPseudoAugment at 10\% run segments (56.7 AP) is almost 10$\times$ more data efficient on the vehicle class, which nearly matches the model trained with 100\% labeled data (57.2 AP). On pedestrian class, AutoPseudoAugment at 10 \% run segments (60.3 AP) shows 5$\times$ data efficient and suprasses no augmentation baseline model trained on 50 \% of the run segments (60.0 AP), shown in \autoref{fig:efficiency}.

\begin{table*}[!h]
\centering
\begin{subtable}[b]{\textwidth}
 \resizebox{\textwidth}{!}{
\begin{tabular}{lll|llllll}
\toprule
\multirow{3}{*}{Setup} & \multirow{3}{*}{Type of data}& \multirow{3}{*}{AutoML} & \multicolumn{6}{c}{Vehicle} \\
 & & & \multicolumn{2}{c}{10 \%} & \multicolumn{2}{c}{30 \%} & \multicolumn{2}{c}{50 \%} \\
 & & & AP (L1/L2) & APH (L1/L2) & AP (L1/L2) & APH (L1/L2) & AP (L1/L2) & APH (L1/L2) \\ \midrule
PPBA \cite{cheng2020improving}& Labeled only & $\checkmark$ &50.2/43.4 & 49.7/42.9 & 56.0/48.7 & 55.5/48.2 &60.9/53.0 & 60.4/52.6\\
Pseudo labeling \cite{caine2021pseudo} & Labeled+Unlabeled & &50.7/43.9 & 50.2/43.5 & 57.8/50.2 & 57.3/49.8 & 59.8/52.0 & 59.3/51.6  \\
AutoPseudoAugment & Labeled+Unlabeled+Fused & $\checkmark$ &\textbf{56.7/49.2} & \textbf{56.3/48.8} & \textbf{61.3/53.5} & \textbf{60.9/53.1} & \textbf{63.0/55.1} & \textbf{62.5/54.6} \\ \bottomrule
\end{tabular}
}
\end{subtable}
\begin{subtable}[b]{\textwidth}
 \resizebox{\textwidth}{!}{
\begin{tabular}{lll|llllll}
\toprule
\multirow{3}{*}{Setup} & \multirow{3}{*}{Type of data}& \multirow{3}{*}{AutoML} & \multicolumn{6}{c}{Pedestrian} \\
 & & & \multicolumn{2}{c}{10 \%} & \multicolumn{2}{c}{30 \%} & \multicolumn{2}{c}{50 \%} \\
 & & & AP (L1/L2) & APH (L1/L2) & AP (L1/L2) & APH (L1/L2) & AP (L1/L2) & APH (L1/L2) \\ \midrule
PPBA \cite{cheng2020improving}& Labeled only & $\checkmark$ &58.5/50.3 & 45.7/39.2 & 61.9/53.7 & 49.4/42.7 & 67.1/58.6 & 54.6/47.5  \\
Pseudo labeling \cite{caine2021pseudo} & Labeled+Unlabeled&   & 56.7/48.5 & 36.7/31.6 & 64.9/56.2 & 48.4/41.8 & 68.2/59.3 & 54.5/47.2 \\
AutoPseudoAugment & Labeled+Unlabeled+Fused & $\checkmark$ &\textbf{60.3/52.1} & \textbf{48.3/41.7} & \textbf{66.5/57.8} & \textbf{55.1/47.7} & \textbf{69.6/60.8} & \textbf{58.9/51.4} \\ \bottomrule
\end{tabular}
}
\end{subtable}
\caption{\textbf{AutoPseudoAugment is more data efficient than SOTA auto data augmentation method (PPBA) and self-training method (Pseudo labeling).} AuotoPseudoAugment outperforms both PPBA and Pseudo labeling when trained on 10\%, 30\%, and 50\% of the labeled training data. For vehicles, with 10\% labeled run segments, AutoPseudoAugment achieves about 6 L1 AP higher compared to PPBA and Pseudo labeling, and matches the quality of 30\% labeled run segments for PPBA and Pseudo labeling. 3D detection Level 1 and 2 detection AP and APH of PointPillars model are evaluated on the Waymo Open Dataset \textit{validation} set.}
\label{tab:AutoPseudoAugment}
\end{table*}

\subsection{Each PseudoAugment is effective.}
Previous sections show the benefit of PseudoAugments are additive to Pseudo labeling and PPBA. In this section, we train PointPillars models on 10\% run segments with only one data augmentation to tease apart the contribution of each PseudoAugment. As a reference, we also show the performance of common data augmentation policies such as random global Z rotation, random global Y rotation, and ground truth bounding box data augmentations \cite{yan2018second,cheng2020improving,ngiam2019starnet,lang2019pointpillars}. Compared to common data augmentation methods, standalone PseudoAugment achieves comparable improvements, shown in \autoref{tab:all_aug}. 

\textbf{PseudoBBox introduces diverse foreground objects.} Unlike using ground truth bounding boxes, PseudoBBox leverages unseen objects in unlabeled data to enrich the training data. On vehicle detection tasks, PseudoBBox significantly outperforms ground truth bounding box (GTBBox) augmentation (+1.7 AP), which highlights the importance of using unseen objects in unlabeled data.

\textbf{PseudoBackground is important.} Interestingly, we observe that utilizing the background point clouds in unlabeled data is important, especially for pedestrian detection. Taking advantage of the unseen backgrounds (PseudoBackground + 3.1 AP) is even more effective to improve model quality compared to using unseen object (PseudoBBox +1.6 AP) for detecting pedestrian.  

\begin{table*}[!t]
\centering
 \resizebox{0.8\textwidth}{!}{
\begin{tabular}{l|l|lll|lll}
\toprule
 & \multirow{2}{*}{No Aug} & \multicolumn{3}{c|}{Common data augmentations} & \multicolumn{3}{c}{PseudoAugments (Ours)} \\
 &  & RotateZ & FlipY & GTBBox & PseudoBBox & PseudoBackground & PseudoFrame \\ \midrule
Vehicle & 41.4 & 45.5 \color[HTML]{3166FF}(+4.1)& 44.4 \color[HTML]{3166FF}(+3.0)& 44.7 \color[HTML]{3166FF}(+3.3)& 46.4 \color[HTML]{3166FF}(+5.0)& 43.0 \color[HTML]{3166FF}(+1.6) &  45.6 \color[HTML]{3166FF}(+4.2)\\
Pedestrian & 49.1 & 52.7 \color[HTML]{3166FF}\color[HTML]{3166FF}(+3.6)& 52.0 \color[HTML]{3166FF}(+2.9)& 50.4 \color[HTML]{3166FF}(+1.3)& 50.3 \color[HTML]{3166FF}(+1.2)& 52.2 \color[HTML]{3166FF}(+3.1) &  49.8 \color[HTML]{3166FF} (+0.7) \\ \bottomrule
\end{tabular}
}
\caption{\textbf{Comparing PseudoAugments with common data augmentations.} PointPillars models are trained with only one data augmentation method on 10\% of the labeled run segments. 3D detection Level 1 AP on Waymo Open Dataset \textit{validation set} are reported. }
\label{tab:all_aug}
\end{table*}

%% file: conclusion.tex
\section{Conclusion}
Despite many prior works on data augmentation for 3D point clouds, data augmentation was mostly based on labeled data.
In this paper, we propose to use unlabeled point clouds to augment training data and introduce PseudoAugments, which utilizes unlabeled point clouds to improve 3D detection. PseudoAugments mitigate intrinsic errors in pseudo labeled scenes while introducing diverse training data by fusing labeled and pseudo labeled scenes. We perform extensive studies and comparisons to show that PseudoAugments generalize to different architectures, model sizes, and datasets and demonstrate that AutoPseudoAugment framework outperforms existing state-of-the-art data augmentation method PPBA \cite{cheng2020improving} and pseudo labeling \cite{caine2021pseudo} at various ratio of labeled and unlabeled data. 

%% file: acknowledgments.tex
\section{Acknowledgments}
We would like to thank Yuning chai, Vijay Vasudevan, Jiquan Ngiam and the rest of Waymo and Google Brain teams for value feedback and infra supports. 